\documentclass[runningheads]{llncs}
\usepackage[T1]{fontenc}
\usepackage{bbding}
\usepackage{graphicx}
\usepackage{makecell}
\usepackage{amsmath}
\usepackage{amssymb}
\begin{document}
\title{Transferring Physical Priors into Remote Sensing Segmentation via Large Language Models}
\titlerunning{Transferring Physical Priors into Remote Sensing Segmentation}
\author{Yuxi Lu\orcidID{0009-0002-2220-7873} \and
Kunqi Li\orcidID{0009-0004-4289-9971} \and
Zhidong Li\orcidID{0000-0002-0784-157X}\and
Xiaohan Su\orcidID{0009-0003-6159-7453}\and
Biao Wu\orcidID{0009-0001-3487-8327}\and
Chenya Huang\orcidID{0009-0008-6686-4393}\and
Bin Liang\Envelope\orcidID{0000-0002-6605-2167}}
\authorrunning{Y. Lu et al.}
\institute{University of Technology Sydney, New South Wales, Australia\\
\email{\{Yuxi.Lu, Kunqi.Li, Xiaohan.Su, Biao.Wu-2, Chenya.Huang\}@student.uts.edu.au}\\ 
\email{\{Zhidong.Li, Bin.Liang\}@uts.edu.au}}
\maketitle              % typeset the header of the contribution
\begin{abstract}
Semantic segmentation of remote sensing imagery is fundamental to Earth observation. Achieving accurate results requires integrating not only optical images but also physical variables such as the Digital Elevation Model (DEM), Synthetic Aperture Radar (SAR) and Normalized Difference Vegetation Index (NDVI). Recent foundation models (FMs) leverage pre-training to exploit these variables but still depend on spatially aligned data and costly retraining when involving new sensors. To overcome these limitations, we introduce a novel paradigm for integrating domain-specific physical priors into segmentation models. We first construct a Physical-Centric Knowledge Graph (PCKG) by prompting large language models to extract physical priors from 1,763 vocabularies, and use it to build a heterogeneous, spatial-aligned dataset, Phy-Sky-SA. Building on this foundation, we develop PriorSeg, a physics-aware residual refinement model trained with a joint visual-physical strategy that incorporates a novel physics-consistency loss. Experiments on heterogeneous settings demonstrate that PriorSeg improves segmentation accuracy and physical plausibility without retraining the FMs. Ablation studies verify the effectiveness of the Phy-Sky-SA dataset, the PCKG, and the physics-consistency loss.

\keywords{Large language models  \and Remote sensing segmentation \and Physical priors.}
\end{abstract}
\section{Introduction}
Semantic segmentation in remote sensing assigns semantic labels to each pixel in satellite or aerial imagery, forming a core task in Earth observation~\cite{yuan2021Review}. It underpins essential applications such as land cover mapping, urban planning, and environmental monitoring~\cite{li2019Deep,wilaiwongsakul2025bare}.

Recent advances in deep learning have driven a shift from low-level texture recognition to high-level semantic understanding, with Foundation Models (FMs) emerging as a representative paradigm. Built upon self-supervised learning, they learn generalized visual representations from large-scale remote sensing data and exhibit strong generalization in downstream tasks such as segmentation~\cite{huang2025survey}.

\begin{figure}[!htbp]
\centering
\includegraphics[width=1\linewidth]{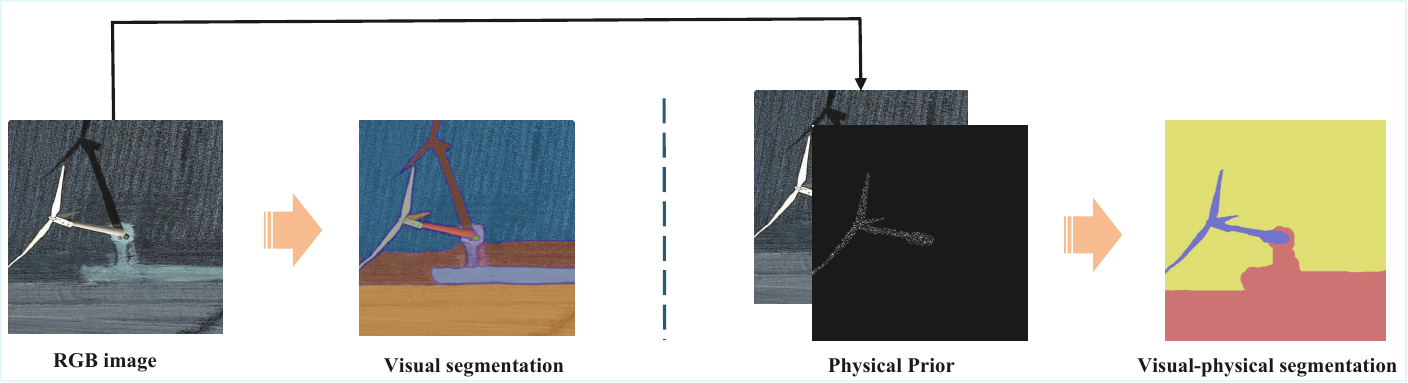}
\caption{Visual and visual-physical segmentation: adding the physical 
prior (simulated SAR) preserves the true object while preventing the shadow from being segmented.}
\label{exa}
\end{figure}

In practice, however, ground objects possess not only visual and textual attributes but also measurable physical variables that are observable by sensors, such as terrain elevation, surface reflectance, and radar backscattering. As illustrated in Fig.\ref{exa}, these physical variables carry essential structural and material properties that complement visual cues and enhance semantic discrimination-particularly in visually ambiguous scenarios. For example, metal roofs and concrete buildings exhibit distinct Synthetic Aperture Radar (SAR) backscatter~\cite{tsokas2022SAR}, flooded~\cite{jia2025comprehensive,jia2025hota} and normal water bodies differ in Digital Elevation Model (DEM) heights
~\cite{guth2021Digital}, and healthy vegetation shows higher Normalized Difference Vegetation Index (NDVI) values than barren or built-up areas~\cite{huang2021Commentary}.

Existing remote sensing FMs (e.g., SkySense~\cite{guo2023SkySense}, RemoteCLIP~\cite{liu2024RemoteCLIP}) integrate spatially aligned physical variables or textual descriptions during pre-training to achieve heterogeneous data alignment via contrastive learning. Despite their strong performance, two major challenges remain: (1) they rely on large, manually aligned datasets, and (2) when involving new sensors or domains, they often require full retraining or architectural modification, limiting scalability and flexibility. This raises a key question: \textit{Can physical priors be efficiently injected into existing FMs without manual spatial alignment or expensive retraining?}

To this end, we propose a novel paradigm for transferring physical priors into segmentation models via large language models (LLMs). We employ LLMs to extract physically plausible numerical ranges from 1,763 vocabularies for each semantic category-covering NDVI, DEM, and SAR-and organize them into a structured \textbf{Physical-Centric Knowledge Graph (PCKG)}. Based on it, we synthesize the heterogeneous dataset \textbf{Phy-Sky-SA}, whose simulated physical variables follow category-level constraints, enabling physically grounded training without manually aligned datasets.

Building on these physical priors, we introduce \textbf{PriorSeg}, a lightweight residual refinement model that learns visual-physical mappings. It attaches to any frozen foundation model and enhances segmentation consistency through structured physical reasoning. At inference, PriorSeg supports two modes: visual-only, producing physically coherent results from learned correlations; and visual-physical, leveraging available variables and the PCKG for further refinement and interpretability. Our contributions are summarized as follows:
\begin{itemize}
    \item A novel LLM-based Physical Prior Transfer paradigm was proposed.
    We transfer physical priors into segmentation models via large language models, which extract domain-specific numerical ranges from 1,763 vocabularies. Based on these extracted priors, we construct the  \textit{PCKG} and \textit{Phy-Sky-SA} dataset, enabling subsequent physically grounded training.
    \item  We design a lightweight model named PriorSeg. It attaches to frozen foundation models and refines their predictions using visual-physical consistency, without retraining the backbone model. It supports both visual-only and visual-physical inference through integrating physical constraints.
    \item Across five datasets and multiple backbones, PriorSeg improves mIoU by \textup{1.14\%--6.7\%}. Ablation studies show that Phy-Sky-SA, the PCKG, and the physics-consistency loss each provide complementary gains, together explaining the full improvement.
\end{itemize}

\section{Methodology}

\subsection{Transferring Physical Priors via LLMs}

\begin{figure}[!htbp]
\centering
\includegraphics[width=1\linewidth]{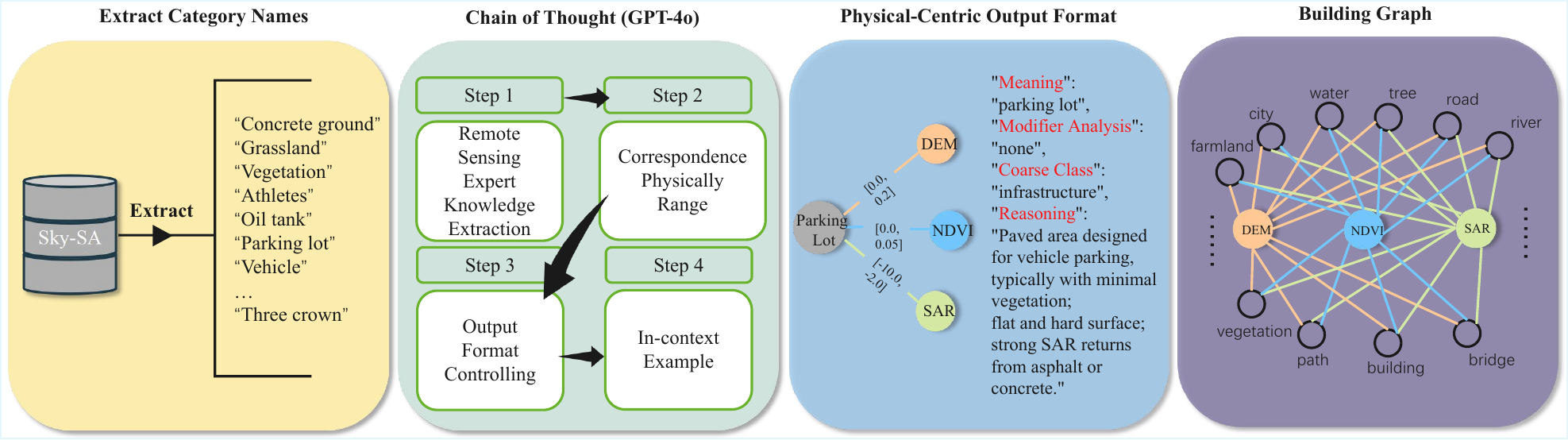}
\caption{Schematic of PCKG construction with GPT-4o.  
Category labels are prompted to the LLM, which returns NDVI, DEM, and SAR
intervals \([a,b]\) plus a brief reasoning trace.  The results are stored
as JSON records and aggregated into a lightweight Physics-Centric
Knowledge Graph that later guides synthetic-modal generation and loss
design.}
\label{3}
\end{figure}

\begin{figure}[!htbp]  
    \centering
    \includegraphics[width=1\linewidth]{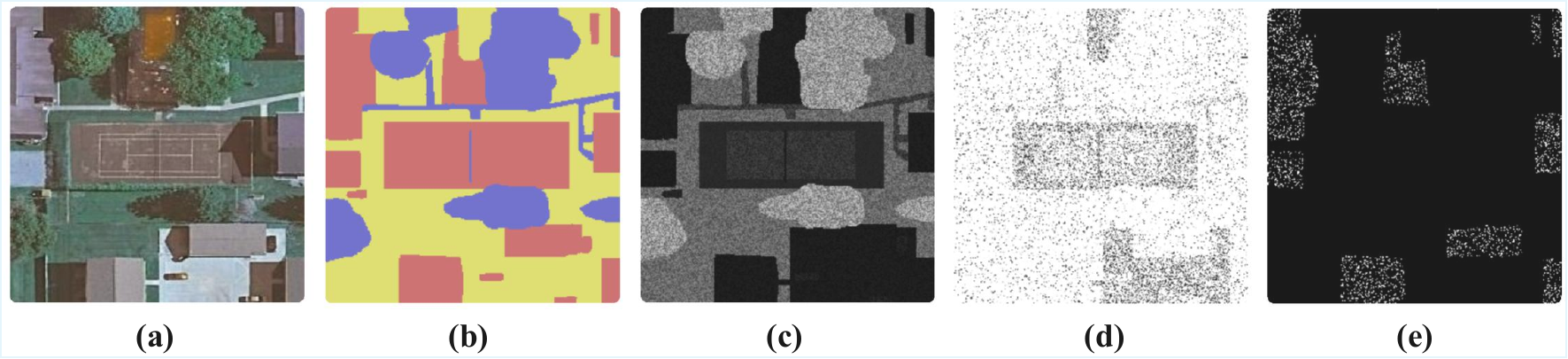}
    \caption{Example sample from the Phy-Sky-SA dataset. From left to right: Image, GT, and the three simulated physical variables-NDVI(S), DEM(S), and SAR(S)-generated
    according to the PCKG, where the “(S)” denotes simulated values.}
    \label{4}
\end{figure}

To incorporate physical priors into segmentation without requiring direct access to manually aligned heterogeneous data, we propose a novel method that leverages LLMs to extract physical priors and inject them into the training process. Our approach comprises two stages:
(1) constructing a structured PCKG, as illustrated in Fig.\ref{3}, and
(2) synthesizing a heterogeneous dataset, Phy-Sky-SA, guided by the PCKG, as shown in Fig.~\ref{4}.

As shown in Fig.\ref{3}, we begin by constructing the PCKG using LLMs. The model is prompted with domain-specific instructions and prior knowledge about three widely used physical variables in remote sensing: DEM\cite{guth2021Digital}, NDVI\cite{huang2021Commentary}, and SAR~\cite{tsokas2022SAR}. These variables are chosen for their broad availability, high discriminative power across land cover types, and general applicability across most semantic categories in remote sensing.

We use the large-scale segmentation dataset \textit{Sky-SA}~\cite{zhuSkySenseO}, which contains 33,776 satellite images and 183,000 mask-text pairs across 1,763 categories, as the textual foundation for category vocabularies. Using its category vocabularies, we design structured prompts that enable LLMs to extract physically plausible value ranges for each semantic class.
Given a category vocabulary (e.g., “bare soil” or “urban park”), the model is asked to:
\begin{itemize}
\item Parse the semantic structure of the category phrase (e.g., target object and its modifiers),
\item Map the phrase to a coarse physical class (e.g., vegetation, road, water),
\item Infer plausible numerical intervals for NDVI, DEM, and SAR values,
\item Justify each inference through step-by-step natural language reasoning.
\end{itemize}

Each entry in the PCKG is stored as a structured JSON object with the following fields: “Category”, “Meaning”, “Modifier Analysis”, “Coarse Class”, “NDVI Range”, “DEM Range”, “SAR Range”, “Reasoning”. The value ranges are defined as closed intervals with two decimal points of precision. Collectively, these entries form a lightweight knowledge base incorporating physical priors for model training and inference.

Having established the PCKG, we integrate it with the Sky-SA dataset to synthesize Phy-Sky-SA, as shown in Fig.~\ref{4}. Although Sky-SA offers rich category diversity, it lacks physical annotations such as DEM, NDVI, and SAR. To address this gap, we generate simulated physical variable maps for each RGB image, guided by the category-specific ranges defined in the PCKG. This process ensures that the synthesized physical maps are pixel-aligned with the original imagery.

The use of LLM-guided prior extraction offers several advantages. It eliminates dependence on manually aligned physical measurements, enables generalization to unseen categories through semantic reasoning. The resulting Phy-Sky-SA dataset then supports training of our lightweight physics-aware refine model PriorSeg.

\subsection{PriorSeg Overview}

\begin{figure*}[t]
    \centering
    \includegraphics[width=1\linewidth]{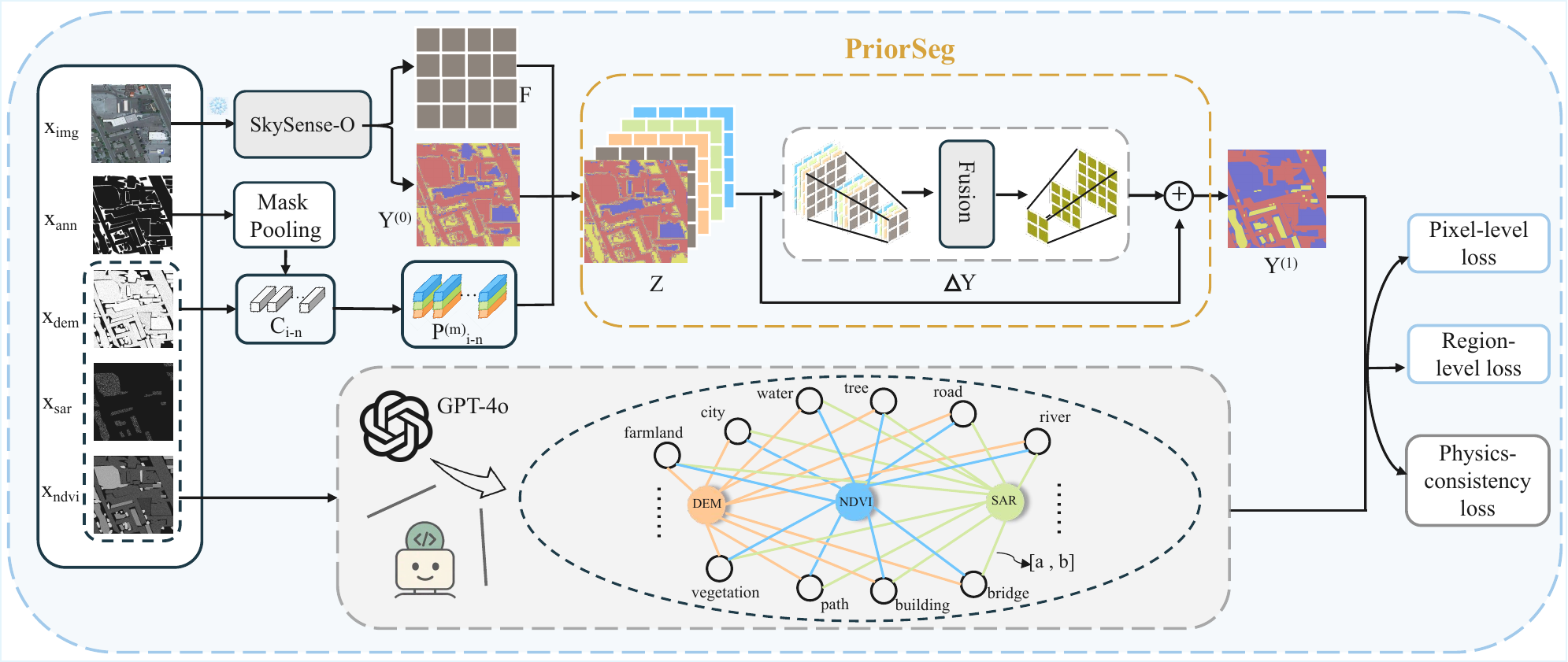}
    \caption{Overview of the PriorSeg training framework. A frozen backbone extracts
    visual features and an initial mask, which are fused with optional physical
    variables. A residual head refines the segmentation, and three losses
    jointly enforce visual quality and physical consistency.}
    \label{1}
\end{figure*}

PriorSeg is a physics-aware residual refinement model that integrates visual
features, coarse predictions, and aligned physical variables into a unified
refinement pipeline (Fig.~\ref{1}). A frozen SkySense-O backbone first produces a visual feature map $F$ 
and an initial coarse mask $Y_{\mathrm{pred}}^{(0)}\!\in[0,1]$ from the RGB image.
Here, the subscript “pred’’ indicates predicted soft segmentation maps.
The aligned physical
rasters $P^{(m)}$ (NDVI, DEM, SAR) are concatenated with $F$ and
$Y_{\mathrm{pred}}^{(0)}$ to form the joint tensor
\[
Z = \bigl[F,\; Y_{\mathrm{pred}}^{(0)},\; P^{(m)}\bigr],
\]
which encodes visual, predictive, and physical cues.

A fusion module processes $Z$, and a residual head predicts a correction 
$\Delta Y$, yielding the refined mask
\[
Y_{\mathrm{pred}}^{(1)} = Y_{\mathrm{pred}}^{(0)} + \Delta Y .
\]
Residual prediction focuses the network on correcting the errors in
$Y_{\mathrm{pred}}^{(0)}$ instead of relearning the mask from scratch, leading
to faster convergence, improved stability, and a lightweight refinement module
that is easily compatible with different frozen backbones.

During training, PriorSeg is supervised by three complementary losses-
pixel-level segmentation, region-level compactness, and physics-consistency-
computed for each class $c$ using the Physical-Centric Knowledge Graph (PCKG),
where $[a,b]$ denotes the admissible physical interval for each class. When
physical variables are unavailable, their channels are zero-padded, enabling
PriorSeg to operate seamlessly in both visual-only and visual-physical modes
using identical model weights.

\subsection{Visual-physical Joint Training}
Training minimises a joint objective on $Y_{\text{pred}}^{(1)}$ alone: a
visual layer (cross-entropy, Dice, region compactness) learns
discriminative features, while a physical-
consistency layer uses a hinge
loss to align each region’s mean physical value with its
PCKG-defined interval, embedding physical priors into the network.
\paragraph{Pixel-level Loss.}
Cross-entropy (\textit{CE}) supplies dense gradients,
while the Dice coefficient rewards spatial overlap and mitigates class
imbalance.
\begin{equation}
L_{\mathrm{seg}}
  = \mathrm{CE}\!\bigl(Y_{\mathrm{pred}}^{(1)},Y_{\mathrm{gt}}\bigr)
  + \alpha\,\mathrm{Dice}\!\bigl(Y_{\mathrm{pred}}^{(1)},Y_{\mathrm{gt}}\bigr),
\end{equation}
where $Y_{\mathrm{gt}}$ is the ground-truth segmentation and $\alpha$ controls the trade-off between the cross-entropy and Dice terms.

\paragraph{Region-level Loss.}
Let $\hat Y=\arg\max Y_{\mathrm{pred}}^{(1)}$ and
$\mathcal{R}_c=\{(x,y)\mid\hat Y(x,y)=c\}$, where $\mathcal{R}_c$ denotes the
set of pixels predicted as class $c$. Define the mean visual feature
$\mu_c = |\mathcal{R}_c|^{-1}\sum_{i\in\mathcal{R}_c}F_i$.
\begin{equation}
L_{\mathrm{region}}
  = \sum_{c=1}^{C}\frac{1}{|\mathcal{R}_c|}
    \sum_{i\in\mathcal{R}_c}\!\lVert F_i-\mu_c\rVert_2^2 ,
\end{equation}
This loss collapses intra-class variance, suppressing speckle noise and
stabilising subsequent physics statistics.

\paragraph{Physics-consistency Loss.}
For each variables $m$, the PCKG provides an admissible interval
$[v_{\min}^{(m,c)}, v_{\max}^{(m,c)}]$, where $c$ indexes the semantic 
classes and $C$ is the total number of classes. 
Denote the mean physical value in region $\mathcal{R}_c$ by
$\bar v_c^{(m)} = |\mathcal{R}_c|^{-1}\sum_{i\in\mathcal{R}_c} v_i^{(m)}$.

\begin{equation}
L_{\mathrm{phys}}^{(m)} =
\sum_{c=1}^{C}
\Bigl[
  \bigl\lvert\bar v_c^{(m)}-v_{\max}^{(m,c)}\bigr\rvert_{+}^{2}
  +\,
  \bigl\lvert v_{\min}^{(m,c)}-\bar v_c^{(m)}\bigr\rvert_{+}^{2}
\Bigr],
\end{equation}
where $|x|_{+}=\max(0,x)$.
Averaging over variables yields
$L_{\mathrm{phys}} = \tfrac1m\sum_{m}L_{\mathrm{phys}}^{(m)}$.
The hinge acts only when $\bar v_c^{(m)}$ departs the legal range,
gently steering predictions toward physical plausibility.

\paragraph{Total Loss.}
The overall objective for visual–physical joint training of \textsc{PriorSeg} is
\begin{equation}
L_{\mathrm{total}}
  = L_{\mathrm{seg}}
  + \lambda_{1}\,L_{\mathrm{region}}
  + \lambda_{2}\,L_{\mathrm{phys}},
\end{equation}
with $\lambda_{1}=0.05$ and $\lambda_{2}=0.40$ tuned on the validation set.

By jointly observing $ Z $ during training, PriorSeg internalises the implicit mapping between visual features and physical rules.

\subsection{Inference}
PriorSeg operates in two modes, both starting from the
\emph{frozen} SkySense-O prediction
$Y_{\text{pred}}^{(0)}$ and feature map $F$.

\paragraph{Visual-only Mode.}
When no physical raster is available, the physics branch of PriorSeg is
gated off.  
The refiner observes $(F, Y_{\text{pred}}^{(0)})$ and predicts a residual
$\Delta Y$, yielding
$Y_{\text{pred}}^{(1)} = Y_{\text{pred}}^{(0)} + \Delta Y$.
The final label is
$\hat Y_i = \arg\max_c Y_{\text{pred}}^{(1)}(i,c)$.
Even without external measurements, PriorSeg relies on its learned
visual-physics memory to correct visually ambiguous regions and produce
physically plausible masks.

\paragraph{Visual-physical Mode.}
If a subset
$\mathcal{M}_{\mathrm{avail}}\subseteq\{\mathrm{NDVI},\mathrm{DEM},\mathrm{SAR}\}$
is provided, the corresponding variables $P^{(m)}$ are supplied to the
physics branch.  
For each pixel $i$, class $c$, and variables $m$, let
$d_i^{(m,c)}$ denote the distance between $v_i^{(m)}$ and its PCKG
interval.  With cap $\tau$ and tolerance $\sigma$,
\begin{equation}
s_i^{(m,c)} =
\exp\!\Bigl[-\min\!\bigl(d_i^{(m,c)},\tau\bigr)^2/\sigma^2\Bigr],
\end{equation}
The attenuation over available variables is
$S_i^{(c)}=\prod_{m\in\mathcal{M}_{\mathrm{avail}}} s_i^{(m,c)}$.
The refiner first produces $\Delta Y$ as in the visual-only path; its
output logits are then re-weighted,
\begin{equation}
\tilde p_i^{(c)} =
\frac{\bigl(Y_{\text{pred}}^{(1)}(i,c)\bigr)\,S_i^{(c)}}
     {\sum_j \bigl(Y_{\text{pred}}^{(1)}(i,j)\bigr)\,S_i^{(j)}},
\quad
\hat Y_i = \arg\max_c \tilde p_i^{(c)}.
\end{equation}
The Gaussian factor $S_i^{(m,c)}$ down-weights classes whose measurements
violate PCKG constraints, while $\tau$ prevents rare outliers from
dominating. If $\mathcal{M}_{\mathrm{avail}}=\varnothing$, then
$S_i^{(c)}\equiv1$ and the procedure degrades gracefully to
visual-only mode.

With real physical variables, PriorSeg yields higher accuracy and stricter
physical consistency; the \textit{reasoning} field in the PCKG provides
explicit interpretability for the applied corrections.

\section{Experiments}
To demonstrate that PriorSeg effectively improves the semantic segmentation accuracy of foundation models, we perform evaluations across five datasets covering diverse scenes and variables combinations, paired with different six backbone models. We observe consistent improvements of approximately 1.14\% to 6.7\% mIoU.

\subsection{Implementation Details}
We train PriorSeg model using two A100 (80\,GB) GPUs. While several advanced LLMs-such as \textit{ChatGPT}~\cite{openai2024GPT4o}, \textit{Gemini}~\cite{comanici2025Gemini}, \textit{LLaMA}~\cite{touvron2023LLaMA}, and \textit{DeepSeek}~\cite{deepseek-ai2024DeepSeek}-offer impressive multi-modal and reasoning capabilities, we adopt \textit{GPT-4o} for its superior performance in grounded reasoning and ease of API-based integration. Training utilizes Sky-SA (SkySense-O), a large-scale remote sensing dataset extended with additional physical variables data using a PCKG generated by GPT-4o.

\begin{table}[t]
\centering

\caption{Performance comparison of six backbone models with and without the proposed PriorSeg refinement module. Results are reported as mIoU(\%) on four datasets: RGB-only of OEM, iSAID, LoveDA, and RGB-Physics WHU-OPT-SAR, SEN12MS.}
\label{tab:backbone_compare}
%\Large
\resizebox{1\linewidth}{!}{%
\setlength{\tabcolsep}{4pt}%
\renewcommand{\arraystretch}{1.15}%
\begin{tabular}{l c c c c c c}
\hline
\textbf{Model} & \textbf{Publication} &
\begin{tabular}{c}RGB\textendash Only\\OEM\end{tabular} &
\begin{tabular}{c}RGB\textendash Only\\iSAID\end{tabular} &
\begin{tabular}{c}RGB\textendash Only\\LoveDA\end{tabular} &
\begin{tabular}{c}RGB\textendash Phy\\WHU-OPT-SAR\end{tabular} &
\begin{tabular}{c}RGB\textendash Phy\\SEN12MS\end{tabular} \\
\hline
SegGPT & ICCV 2023 & 25.19 & 29.17 & 31.40 & 23.42 & 21.84 \\
\textbf{SegGPT+PriorSeg} &  & \textbf{27.33(+2.14)} & \textbf{32.18(+3.01)} & \textbf{33.55(+2.15)} & \textbf{29.33(+5.91)} & \textbf{27.56(+5.72)}\\
DINOv & CVPR 2023 & 23.05 & 25.35 & 25.17 & 22.74 & 20.09 \\
\textbf{DINOv+PriorSeg} &  & \textbf{24.32(+1.27)} & \textbf{27.39(+2.04)} & \textbf{27.95(+2.78)} & \textbf{28.67(+5.93)} & \textbf{25.44(+5.35)}\\
SAN & CVPR 2023 & 24.87 & 11.77 & 27.83 & 25.97 & 21.33 \\
\textbf{SAN+PriorSeg} &  & \textbf{27.22(+2.35)} & \textbf{15.24(+3.47)} & \textbf{30.51(+2.68)} & \textbf{32.71(+6.74)} & \textbf{26.76(+5.43)}\\
CAT-SEG & CVPR 2024  & 30.26 & 20.58 & 37.15 & 35.22 & 32.17\\
\textbf{CAT-SEG+PriorSeg} &  & \textbf{32.38(+2.12)} & \textbf{23.55(+2.97)} & \textbf{39.04(+1.89)} & \textbf{39.47(+4.25)} & \textbf{37.16(+4.99)}\\
SegEarth-OV & Arxiv 2024  & 40.30 & 21.70 & 36.90 & 31.61 & 30.21\\
\textbf{SegEarth-OV+PriorSeg} &   & \textbf{41.98(+1.68)} & \textbf{22.85(+1.15)} & \textbf{38.37(+1.47)} & \textbf{35.74(+4.13)} & \textbf{34.69(+4.48)}\\
SkySense-O & CVPR 2025  & 40.83 & 43.92 & 38.30 & 39.83 & 39.71\\
\textbf{SkySense-O+PriorSeg} &  & \textbf{41.97(+1.14)} & \textbf{45.23(+1.31)} & \textbf{39.72(+1.42)} & \textbf{42.98(+3.15)}& \textbf{43.10(+3.39)} \\
\hline
\end{tabular}%
}
\end{table}

\begin{figure}[t]  
    \centering
    \includegraphics[width=0.9\columnwidth,keepaspectratio]{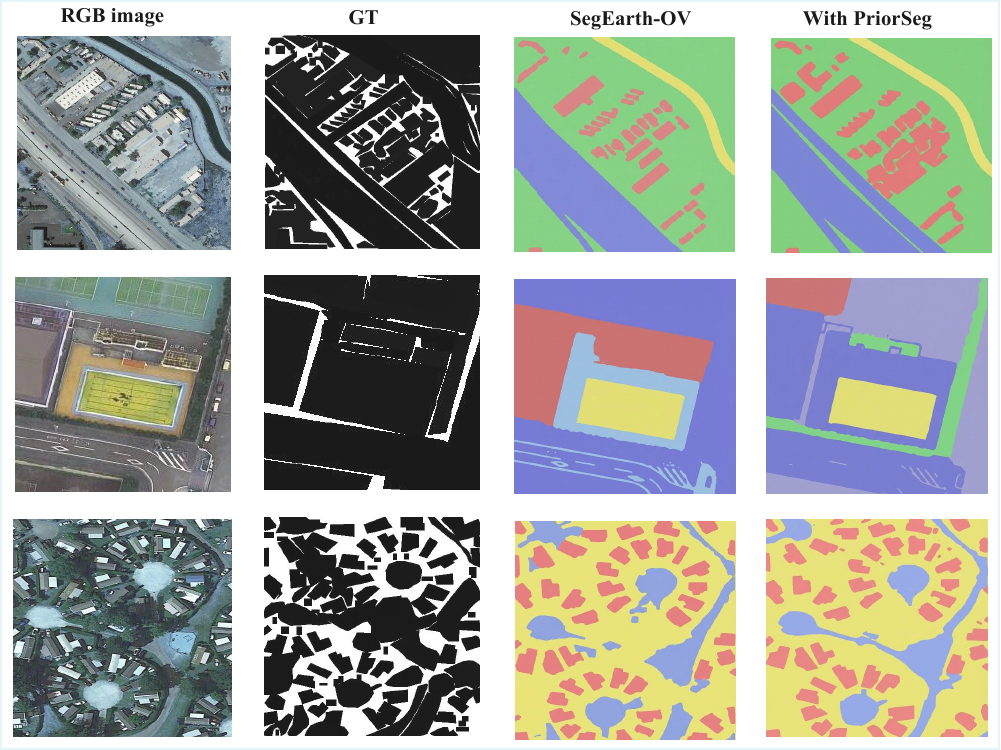}
    \caption{Three representative visualization examples on SegEarth-OV backbone, comparing GT, SegEarth-OV predictions, and our PriorSeg-enhanced results.}
    \label{6}
\end{figure}

\subsection{Zero-Shot Refinement Evaluation}
We evaluate PriorSeg in two settings. In the \emph{visual-only} mode, the model is tested on three external RGB benchmarks-iSAID~\cite{waqas2019isaid}, OEM~\cite{wu2013oem}, and LoveDA~\cite{wang2021loveda}-without any fine-tuning or exposure to their labels. Even without physical variables, PriorSeg consistently improves mIoU across all backbones (Table~1), indicating that its visual–physics memory generalizes from vision alone. In the \emph{visual-physical} mode, PriorSeg is evaluated on WHU-OPT-SAR and SEN12MS, which provide real NDVI and SAR. Without retraining, it achieves the largest gains as physical variables activate PCKG-guided reasoning. As shown in Fig.~\ref{6}, representative examples visually confirm these refinements.

\subsection{Evaluation with Different Backbones}
Because PriorSeg operates purely on the feature map and first-pass mask of a frozen backbone model, it can attach to arbitrary architectures.  We test six backbones-SegGPT~\cite{wang2023seggpt1}, DINOv~\cite{li2024visual}, SAN~\cite{Xu_2023_CVPR}, CAT-Seg~\cite{cho2024cat}, SegEarth-OV~\cite{li2025segearth}, and the original SkySense-O~\cite{zhuSkySenseO}.  When a backbone’s output visual feature shape differs from that of SkySense-O, a lightweight \(1{\times}1\) adapter is inserted to match the expected tensor size; only the refiner (and, if present, the adapter) is trained. Across the four external datasets PriorSeg delivers consistent mIoU gains on every backbone (Table.~1), confirming that the proposed refiner generalises well beyond its source model. The residual refiner is conservative with respect to already correct regions. When applied to a strong baseline such as SkySense-O, which has been extensively trained on the Sky-SA dataset, PriorSeg performs only subtle adjustments on pixels whose physical readings violate the PCKG intervals or whose confidence is low. As a result, the observed gains are concentrated in areas exhibiting pronounced physical discrepancies, and the overall improvement remains moderate.

\begin{figure}[t]  
    \centering
    \includegraphics[width=0.7\columnwidth,keepaspectratio]{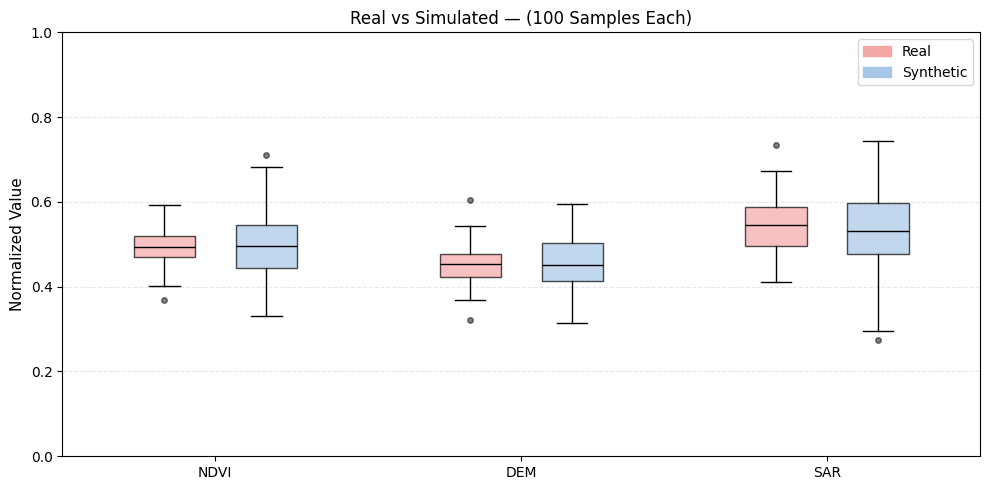}
    \caption{Comparison between real and simulated physical variables on the WHU-OPT-SAR dataset.}
    \label{5}
\end{figure}

\subsection{Reliability Analysis}
To assess the reliability of simulated physical variables, we replicate the Phy-Sky-SA synthesis procedure on the WHU dataset, generating SAR, DEM, and NDVI variables from ground truth labels and PCKG intervals. The NDVI is computed as $(\text{NIR}-\text{Red})/(\text{NIR}+\text{Red})$, while DEM data are sourced from manually preprocessed SRTM (30\,m resolution) tiles covering Hubei Province (30--33\textdegree N, 108--117\textdegree E). 

Fig.~\ref{5} shows strong distributional overlap across NDVI, DEM, and SAR. NDVI matches most closely, DEM shows moderate variation, and SAR displays a wider natural spread. Median alignment and compact ranges confirm that PCKG-guided synthesis produces realistic physical characteristics suitable for downstream applications.

\begin{table}[t]
\centering
\caption{The ablation results of the SAN model on the WHU-OPT-SAR dataset, evaluating Phy-Sky-SA, PCKG, and $\mathcal{L}_{\text{phys}}$.}
\setlength{\tabcolsep}{6pt}
\renewcommand{\arraystretch}{1.15}
\begin{tabular}{ccc|c}
\hline
\textbf{+Phy-Sky-SA} & \textbf{+PCKG} & \textbf{+$\mathcal{L}_{\text{phys}}$} & \textbf{mIoU (\%)} \\
\hline
             &               &               & 25.97 \\
$\checkmark$ &               &               & 27.69\textbf{(+1.72)} \\
$\checkmark$ & $\checkmark$  &               & 29.93\textbf{(+2.24)} \\
$\checkmark$ & $\checkmark$  & $\checkmark$  & 32.71\textbf{(+2.78)}\\
\hline
\end{tabular}
\label{2}
\end{table}

\subsection{Ablation Study}
Table~\ref{2} evaluates the contributions of Phy-Sky-SA, PCKG, and physics-consistency loss. Training on Phy-Sky-SA yields a 1.72\% gain. Introducing the PCKG adds 2.24\% by injecting pixel-wise physical logic. Adding the physics-consistency loss provides another 2.78\%, penalizing visually ambiguous yet physically implausible predictions. Together, these components enable PriorSeg to leverage joint visual–physical information and deliver coherent refinement results.

\section{Conclusion}
We introduced a paradigm for transferring physical priors into remote sensing segmentation without requiring manually aligned datasets or retraining foundation models. Using large language models, we constructed the PCKG and the Phy-Sky-SA dataset with physically constrained NDVI, DEM, and SAR variables. Built on these priors, PriorSeg serves as a lightweight refinement module that enforces visual-physical consistency for frozen backbones. Experiments show improved segmentation accuracy and physical plausibility, and future work will incorporate more types of physical variables.

\bibliographystyle{splncs04}

\bibliography{refs}

\end{document}